# ASI: Accuracy-Stability Index for Evaluating Deep Learning Models

Wei Dai
Department of Computer Science
Purdue University Northwest
Hammond, Indiana, USA
weidai@pnw.edu

Daniel Berleant
Department of Information Science
Univeristy of Arkansas at Little Rock
Little Rock, Arkansas, USA
jdberleant@ualr.edu

*Abstract*— In the context of deep learning research, where model introductions continually occur, the need for effective and efficient evaluation remains paramount. Existing methods often emphasize accuracy metrics, overlooking stability. To address this, the paper introduces the Accuracy-Stability Index (ASI), a quantitative measure incorporating both accuracy and stability for assessing deep learning models. Experimental results demonstrate the application of ASI, and a 3D surface model is presented for visualizing ASI, mean accuracy, and coefficient of variation. This paper addresses the important issue of quantitative benchmarking metrics for deep learning models, providing a new approach for accurately evaluating accuracy and stability of deep learning models. The paper concludes with discussions on potential weaknesses and outlines future research directions.

## I. INTRODUCTION

Benchmarking is essential for a variety of reasons in everyday life, industry, and science. People choose benchmarking tools to evaluate products, quality, and values. In the computing industry, major benchmarking organizations include: the Performance Evaluation Corporation (SPEC) [1][2], which measures CPUs; the Transaction Processing Performance Council (TPC) [3][4], which evaluates databases; and the Storage Performance Council (SPC) [4], which audits disk storages. Additionally, the Machine Learning Performance (MLPerf) [5] organization evaluates deep learning hardware [6].

Quantitative benchmarking is indispensable in the software industry, offering technical guidelines that aid engineers and scientists in accurately evaluating software and hardware. These metrics establish standards, providing numerical indexes to discern the quality of software and algorithms. In the field of deep learning research, both industrial and academic researchers continually introduce new models. Despite this influx, the ability to effectively and efficiently evaluate these models remains an unfinished problem. Hence, quantitative benchmarking advances hold critical importance in the realm of deep learning research.

Existing quantitative benchmark methods for deep learning models are straightforward, often involving the assessment of Top-n accuracy metrics, such as Top-1, Top-3, and Top-5. In the context of these metrics, Top-1 accuracy signifies that the deep learning model's response with the highest confidence precisely matches the ground truth. On the other hand, Top-5 accuracy requires any of the top 5 highest confidence results from the model to exactly match the ground truth. However, achieving high accuracy does not guarantee stability or robustness[7]. Critically, pursuing accuracy may lead to overfitting in deep learning models, where accurate predictions are made during the training stage but not necessarily for new data later.

Current research typically measures the Top-n accuracy of deep learning models, assuming that all input data during both the training and inference stages are high-quality datasets, such as pristine images. However, in the real world, the quality of datasets often varies. For instance, images may become blurred by cameras installed in vibrating environments, and image quality can be compromised when strong lights interfere with the camera lens[8].

Unfortunately, existing research often emphasizes the accuracy of deep learning models while overlooking their stability. Studies indicate that most deep learning models demonstrate instability when presented with low-quality images, leading to substantial reductions in accuracy. Clearly, the goal should not be to have deep learning models that only provide either high accuracy or high stability. Instead, there is a need for deep learning models that can achieve both high accuracy and high stability. The relationship between accuracy and stability is illustrated in Fig. 1.

To address the problems stated above, our research aim is to develop a quantitative index for assessing deep learning models that integrate two key metrics: accuracy and stability.

In this paper, our contributions are as follows:

- We show that the robustness of deep learning shares the same mathematical formula as robustness to adversarial attacks, though their optimization approaches differ.

- We introduce the **A**ccuracy-**S**tability **I**ndex (ASI) for measuring both accuracy and stability of deep learning models. Additionally, we conducted experimental tests, demonstrating how to evaluate deep learning models using ASI.

- We provide a 3D ssurface model for visualizing ASI, mean accuracy, and coefficient of variation.

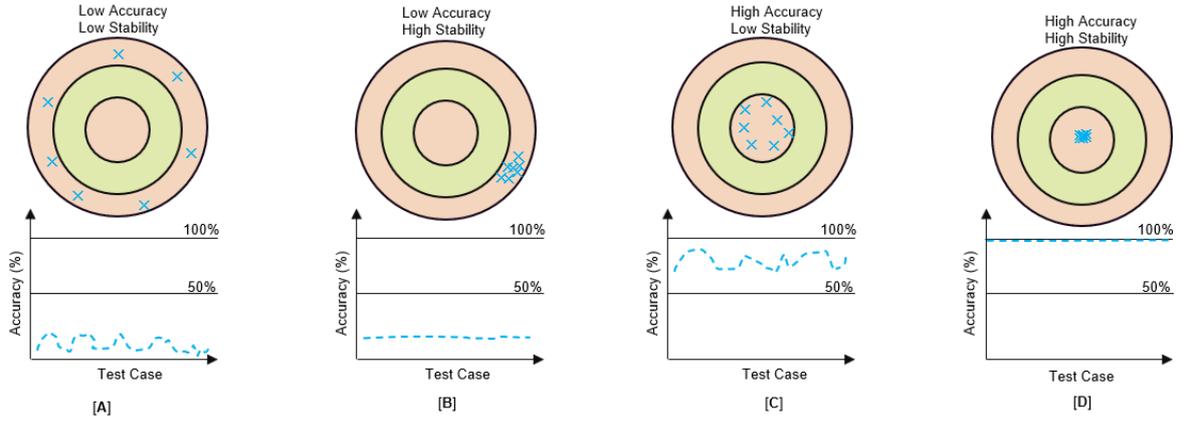

**Figure 1.** Pictorial representation of the relationship between accuracy and stability

The remainder of this paper is organized as follows. Section II is Related Work. In Section III, we provide the research design. Section IV introduces the experimental results. Discussion, Conclusion, and Future Work constitute Section V, Section VI, and Section VII, respectively

## II. RELATED WORK

Image perturbations impact image quality assessment and accuracy of deep learning models. In [9], the authors report that BRISQUE and PSNR, two major image quality assessments, do not work well on perturbed image sets. Researchers have uncovered the instability of deep learning models to image perturbations. In [7], it was demonstrated that the Google Cloud Vision API could be deceived with an addition of approximately 14.25% impulse noise density. Accuracy was reduced in deep learning models when images were subjected to various quality distortions, such as JPEG/JPEG2000 compression, blur, Gaussian noise, and contrast [10]. Thus, the formula is expressed as Eq. 1:

$$f(x + \delta) \not\equiv f(x) \quad (1)$$
$$subject\ to \quad x + \delta \in R\ and\ \delta \neq 0$$

where $f(.)$ is a deep learning model, $x$ codes an original image, and $\delta$ codes an image perturbation.

Previous studies have reported corruptions imposed on high quality ImageNet datasets, see Fig. 2. In [11], the authors introduced 15 types of perturbations, including Gaussian noise, blur, and brightness. While the investigation generated flawed images through single-factor corruptions, real-world scenarios often involve the impact of more than one perturbation on image quality. For instance, adverse weather conditions (e.g, fog) and vibrations could simultaneously affect the cameras of self-driven cars, leading to unforeseen traffic accidents. In [8], the authors devised 69 benchmarking image sets, comprising a clean set, sets with single-factor perturbation conditions, and sets with two-factor perturbation conditions, to assess the robustness of deep learning models.

Apart from unintentional image noises, adversarial attacks represent another challenge to the accuracy of deep learning models. In the context of adversarial attacks, δ represents specially designed noise, where $f(x+\delta)$ corresponds to the target label. Many adversarial attack methodologies incorporate optimizations to streamline computational complexity and reduce processing time. An adversarial image is a high quality picture intentionally altered with subtle corruptions specifically chosen to bewilder deep learning models. While these corruptions might go unnoticed by humans, research indicates that adversarial examples of deliberately modified images can deceive deep learning models, resulting in confused classifiers and diminished overall accuracy of the models [12]–[16].

Let us summarize the differences between evaluating deep learning robustness to ordinary noise and to adversarial attacks. For details, see Table I.

TABLE I. DIFFERENCE BETWEEN ORDINARY NOISE AND ADVERSARIAL ATTACKS

| Item | Ordinary Noise | Adversarial Attacks |
|---|---|---|
| Perturbation for Human Eyes | Natural noise, easily identified | Small amounts of noise, difficult to identify. |
| Label is Targeted | No | Yes |
| Computing Complexity | Low | High |
| Speed | Fast | Slow |
| Generalizability | High | Low |
| Motivation | Positive | Negative |

Benchmarking metrics find diverse application, with previous researchers assessing the robustness of deep learning (DL) classifiers through Top-1 and Top-5 precision analyses [17]. For instance, [11] employed the average rates of correct ("Top-1") and almost correct ("Top-5") classifications, with the Top-1 error rate of AlexNet serving as the reference. In [8] a two-dimensional metric was introduced comprising mean accuracy and coefficient of variation (CV). This approach aids in characterizing the stability of DL classifiers by emphasizing high mean accuracy and a small CV across various perturbation conditions including two-factor perturbations (Table II) to better reflect the messiness of real world applications.

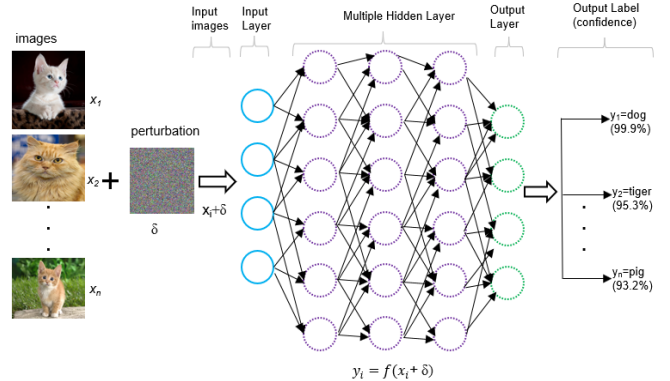

**Figure 2.** Unstable Deep Learning Models

TABLE II. TWO-FACTOR PERTURBATION SEQUENCES, TYPES, AND INTENSITIES

| Perturbation Sequence | Respective Noise Intensities Tested for Perturbations 1 & 2 | |
|---|---|---|
| SP-Gaussian | 0.1, 0.15, & 0.2 | 0.1, 0.15, & 0.2 |
| Gaussian-SP | 0.1, 0.15, & 0.2 | 0.1, 0.15, & 0.2 |
| SP-Rotation | 0.1, 0.15, & 0.2 | -60°, -30°, 0°, 30°, & 60° |
| Rotation-SP | -60°, -30°, 0°, 30°, & 60° | 0.1, 0.15, & 0.2 |

A full discussion appears in [8] and [9].

## III. RESEARCH DESIGN

The research employed images intentionally crafted to exhibit defects through two-factor perturbation [8]. The formula is written as Eq. 2:

$$f(x + \delta) \not\equiv f(x) \quad (2)$$
$$\text{subject to } \delta = \alpha \oplus \beta, \quad \alpha \oplus \beta \neq \beta \oplus \alpha$$

where $f(.)$ is a deep learning model, $x$ is an original image, and $\alpha$ and $\beta$ are image perturbations. Note that $\alpha \oplus \beta \neq \beta \oplus \alpha$, in general, because perturbations are non-commutative, as the order that perturbations are applied matters.

### A. Two-Factor Image Perturbations

Two-factor perturbation involves the alteration of images using two distinct types of perturbation consecutively. These dual perturbations might encompass digital perturbations such as salt & pepper noise and Gaussian noise applied in different sequences. Another scenario of two-factor perturbation involves a geometric perturbation, such as rotation, followed by a noise perturbation, or vice versa, with noise preceding the rotation. For details, see Table II.

### B. Benchmarking Formula

The **c**oefficient of **v**ariation (CV) is defined as the ratio of the standard deviation to the mean as shown in Eq. 3. The CV is a statistical metric that quantifies the relative variability within a set of data points. A reduced CV implies diminished relative variability, indicating a more uniform set of values. Conversely, an elevated CV suggests heightened relative variability or dispersion within the data.

$$CV\ (\%) = 100 \times \frac{Standard\ Deviation}{Mean} \quad (3)$$

Each testing set comprised 68 corrupted image groups and one pristine image group. The clean image group contained the original, unaltered images, and each group consisted of 500 images. There is no overlap between the training sets and testing sets. A DL classifier trained on clean (uncorrupted) data is notated as $D_{testset}$ where $D$ refers to the DL algorithm and *testset* refers to a test set. Then, we tested the accuracy on each corruption type $c$. Different $c$ values are different corruption types, and level of severity is $s$, with $s \in \{0.1, 0.15, 0.2\}$ when doing SP or GA corruptions, and a $s \in \{-60°, -30°, 0°, 30°, 60°\}$ when doing rotation corruptions. The mean accuracy is written $Accu(D)$ as shown in Eq. 4.

$$Accu(D)\ \% = 100 \times \frac{\sum_{c=0}^{68} Accu(D_{testset_c})}{69} \quad (4)$$
$$c \in [0, 68]$$

where $Accu(D_{testset_c})$ refers to the specific accuracy with respect to test set $c$. The range of possible values for $c$ spans from 0 to 68, as specified in [8][9].

$$CV(D)\% = 100 \times \frac{\sqrt{\frac{\sum_{c=0}^{68}(Accu(D_{testset_c}) - Accu(D))^2}{69}}}{Accu(D)} \quad (5)$$

Here, $CV(D)$ stands for the coefficient of variation of the mean of the accuracies of deep learning model $D$.

We introduce a benchmarking index to quantify the mean accuracies and coefficients of variation (CVs) of DL classifiers. The index is computed as the ratio of the difference between mean accuracy and CV over their sum. This normalization brings the ratio within the range of [-1, 1]. Let us name it the Accuracy-Stability Index ("ASI" for short).

$$ASI = \frac{Accu(D) - CV(D)}{Accu(D) + CV(D)} \quad (6)$$

subject to $Accu(D) + CV(D) \neq 0$

where $Accu(D)$ and $CV(D)$ were defined by Eqs. 3-5.

In Eq. 6, this normalization provides a standardized measure that enables the assessment of the interplay between mean accuracy and dataset stability in deep learning models. A higher ASI value indicates a more favorable balance between mean accuracy and stability, while a lower ASI suggests a less desirable balance.

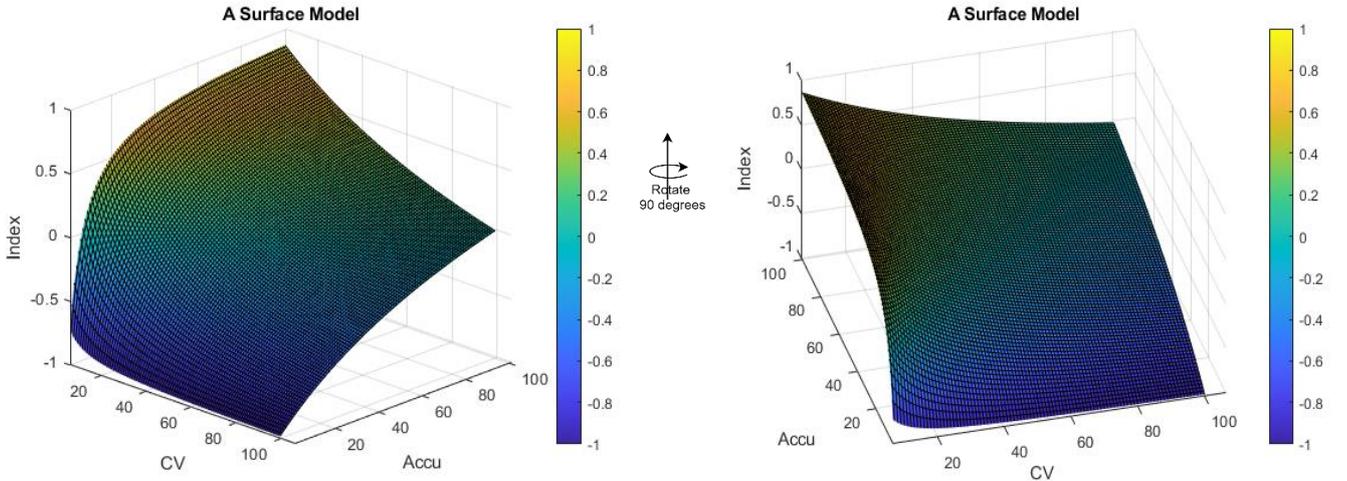

**Figure 3.** 3D Surface model for Accuracy-Stability Index.

Note that the left and right figures show the same 3D surface model, but from different directions.

## IV. Experimental Results

Comparing row 1 (R1 for short, and so on) to R2 in Table III, the CV of R1 is greater than for R2, but the mean of accuracies is less. It is straightforward therefor that the robustness of R2 is better than R1. Comparing the ASI, the ASI of R2 is 0.973, larger than the corresponding ASI for R1 of 0.948. Next, let us compare R8 to R4. The CV of R4 is 1.479, but the CV of R8 is 1.737. The CV of R8 is 17.442% above that of R4 since $(\frac{1.737}{1.479} - 1) \times 100\% = 17.442\%$. The mean of accuracies of R8 is -1.158% below that of R4, since $(\frac{88.663}{89.702} - 1) \times 100\% = -1.158$. It is clear that R4 is better than R8 because R4 has a smaller CV and larger mean accuracy. Correspondingly the ASI of R4, 0.968, is greater than that of R8, 0.963.

## V. Discussion

The ASI benchmarking index is defined as a fraction, the difference between the mean accuracy and the coefficient of variation (CV) over their sum (Eq. 6, simplified in Eq. 7).

$$ASI = \frac{Mean\ Accuracy\ - CV}{Mean\ Accuracy\ + CV} \quad (7)$$

This formula computes a ratio, normalized to within [-1, 1], which standardizes the measure and makes it more interpretable. The resulting ASI metric can help assess the accuracy and stability of deep learning classifiers by considering both their mean accuracies and their variation in performance across different datasets. If the ASI is closer to 1, it indicates a more favorable balance between mean accuracy and stability; if closer to -1, it suggests a less desirable balance. Thus, the ASI metric quantifies an integrated composite of the accuracy and variation performances of deep learning models.

The ASI has two variables, mean accuracy and CV. For a visual depiction of the relationship between accuracy and stability, see Fig. 3. Let us discuss potential weakness of the benchmark metrics. Firstly, mean accuracy is sensitive to outliers. If there are extreme lower/higher accuracy of deep learning models under special conditions, mean accuracy should be impacted, leading to a potentially skewed representation of the overall performance.

Secondly, CV makes the common assumption that the data are normally distributed. If the data have different underlying distributions, the CV would be less dependable as a measure of relative variability.

## VI. Conclusion

The paper introduced the Accuracy-Stability Index (ASI) as a novel quantitative measure, incorporating both accuracy and stability, specifically designed for assessing deep learning models. Experimental results demonstrated the application of ASI, and a 3D surface plot was presented (Fig. 3) for visualizing its behavior. The paper concluded with discussions on potential weaknesses and outlines, next, directions for future research, including consideration of the sensitivity of the ASI metric to small fluctuations.

## VII. Future Work

Currently, our experimental results show that ASI works well as intended. However, ASI may be sensitive to small changes in either Mean Accuracy or CV, potentially causing larger variations in the index. Sensitivity to small fluctuations could lead to instability in the interpretation of the index. To investigate this further, the ASI metric (Eqs. 6 & 7) should be compared to the harmonic mean, or F1 measure that is classically popular in the information retrieval field and more recently in other fields in particular ML. Another direction for future work suggested by reviewers of this article is to investigate the ASI metric on more diverse datasets, not only different types of images, but also audio files or other complex non-image data sets. Additionally, new and better deep learning models are expected to be introduced in the coming years and it would be useful to evaluate these future models on their performance on the ASI metric, which could contribute a valuable new tool in the deep learning model assessment and benchmarking field.

## Acknowledgment

The project was partially supported by a Google grant. We are grateful for the penetrating yet positive reviewer insights and comments on future work and the potential to contribute to the development of the deep learning field.

## References

[1] J. Bucek, K.-D. Lange, *et al.*, "SPEC CPU2017: next-generation compute benchmark," in *Companion of the 2018 ACM/SPEC International Conference on Performance Engineering*, 2018, pp. 41–42.

[2] J. L. Henning, "SPEC CPU2000: measuring CPU performance in the new millennium," *Computer* (Long Beach Calif), 2000, doi: 10.1109/2.869367.

[3] M. Poess, "TPC's benchmark development model: making the first industry standard benchmark on big data a success," in *Specifying Big Data Benchmarks*, Springer, 2014, pp. 1–10.

[4] Storage Performance Corporation, SPC. [Online]. Available: http://www.spcresults.org/

[5] P. Mattson, "MLPerf training algorithms." [Online]. Available: https://github.com/mlperf/training

[6] W. Dai and D. Berleant, "Benchmarking contemporary deep learning hardware and frameworks: a survey of qualitative metrics," *2019 IEEE First International Conference on Cognitive Machine Intelligence (CogMI)*, pp. 148–155, Dec. 2019, doi: 10.1109/CogMI48466.2019.00029.

[7] H. Hosseini, B. Xiao, and R. Poovendran, "Google's cloud vision API is not robust to noise," in *Proceedings - 16th IEEE International Conference on Machine Learning and Applications, ICMLA 2017*, 2018. doi: 10.1109/ICMLA.2017.0-172.

[8] W. Dai and D. Berleant, "Benchmarking robustness of deep learning classifiers using two-factor perturbation," in *Proceedings - 2021 IEEE International Conference on Big Data, Big Data 2021*, 2021. doi: 10.1109/BigData52589.2021.9671976.

[9] W. Dai and D. Berleant, "Discovering limitations of image quality assessments with noised deep learning image sets," in *Proceedings - 2022 IEEE International Conference on Big Data, Big Data 2022*, 2022. doi: 10.1109/BigData55660.2022.10020507.

[10] S. Dodge and L. Karam, "Understanding how image quality affects deep neural networks," in *2016 Eighth International Conference on Quality of Multimedia Experience (QoMEX)*, 2016, pp. 1–6.

[11] D. Hendrycks and T. G. Dietterich, "Benchmarking neural network robustness to common corruptions and perturbations," *arXiv Preprint, arXiv:1903.12261*, 2019.

[12] R. Geirhos, D. H. J. Janssen, H. H. Schütt, J. Rauber, M. Bethge, and F. A. Wichmann, "Comparing deep neural networks against humans: object recognition when the signal gets weaker," *arXiv Preprint, arXiv:1706.06969*, 2017.

[13] M.-I. Nicolae, *et al.*, "Adversarial Robustness Toolbox v1.0.0," *arXiv Preprint, arXiv:1807.01069*, 2018.

[14] A. Kurakin, I. Goodfellow, and S. Bengio, "Adversarial machine learning at scale," *arXiv Preprint, arXiv:1611.01236*, 2016.

[15] A. Madry, A. Makelov, L. Schmidt, D. Tsipras, and A. Vladu, "Towards deep learning models resistant to adversarial attacks," *arXiv Preprint, arXiv:1706.06083*, 2017.

[16] N. Carlini and D. Wagner, "Adversarial examples are not easily detected: Bypassing ten detection methods," in *Proceedings of the 10th ACM Workshop on Artificial Intelligence and Security*, 2017, pp. 3–14.

[17] S. Zheng, Y. Song, T. Leung, and I. Goodfellow, "Improving the robustness of deep neural networks via stability training," in *Proceedings of the IEEE Computer Society Conference on Computer Vision and Pattern Recognition*, 2016. doi: 10.1109/CVPR.2016.485.

TABLE III. IMAGE CORRUPTION CONDITIONS AND ASI.

| Row ID | Condition | DL classifier | CV (%) | Mean of accuracies (%) | ASI |
|---|---|---|---|---|---|
| 1 | (*) clean | AlexNet | 2.276 | 85.250 | 0.948 |
| 2 | 5%_noise | AlexNet_5P_SP0.1GA0.1 | 1.246 | 89.970 | 0.973 |
| 3 | 10%_noise | AlexNet_10P_SP0.1GA0.1 | 1.167 | 89.966 | 0.974 |
| 4 | 15%_noise | AlexNet_15P_SP0.1GA0.1 | 1.479 | 89.702 | 0.968 |
| 5 | 5%_noise | AlexNet_5P_GA0.1SP0.1 | 0.598 | 89.809 | 0.987 |
| 6 | 10%_noise | AlexNet_10P_GA0.1SP0.1 | 0.948 | 89.836 | 0.979 |
| 7 | 15%_noise | AlexNet_15P_GA0.1SP0.1 | 1.395 | 89.477 | 0.969 |
| 8 | 5%_noise | AlexNet_5P_SP0.1RR30 | 1.737 | 88.663 | 0.962 |
| 9 | 10%_noise | AlexNet_10P_SP0.1RR30 | 2.252 | 87.328 | 0.950 |
| 10 | 15%_noise | AlexNet_15P_SP0.1RR30 | 3.331 | 85.663 | 0.925 |
| 11 | 5%_noise | AlexNet_5P_SP0.1RL30 | 2.631 | 87.723 | 0.942 |
| 12 | 10%_noise | AlexNet_10P_SP0.1RL30 | 2.693 | 86.218 | 0.939 |
| 13 | 15%_noise | AlexNet_15P_SP0.1RL30 | 3.272 | 86.021 | 0.927 |
| 14 | 5%_noise | AlexNet_5P_SP0.1GA0 | 0.786 | 89.807 | 0.983 |
| 15 | 10%_noise | AlexNet_10P_SP0.1GA0 | 1.014 | 89.865 | 0.978 |
| 16 | 15%_noise | AlexNet_15P_SP0.1GA0 | 1.512 | 89.245 | 0.967 |
| 17 | 5%_noise | AlexNet_5P_GA0.1SP0 | 0.465 | 89.801 | 0.990 |
| 18 | 10%_noise | AlexNet_10P_GA0.1SP0 | 1.342 | 89.932 | 0.971 |
| 19 | 15%_noise | AlexNet_15P_GA0.1SP0 | 1.367 | 89.306 | 0.970 |
| 20 | 5%_noise | AlexNet_5P_RL30SP0 | 2.998 | 86.107 | 0.933 |
| 21 | 10%_noise | AlexNet_10P_RL30SP0 | 2.960 | 84.935 | 0.933 |
| 22 | 15%_noise | AlexNet_15P_RL30SP0 | 3.733 | 84.453 | 0.915 |
| 23 | 5%_noise | AlexNet_5P_RR30SP0 | 2.243 | 87.504 | 0.950 |
| 24 | 10%_noise | AlexNet_10P_RR30SP0 | 2.578 | 86.142 | 0.942 |
| 25 | 15%_noise | AlexNet_15P_RR30SP0 | 3.536 | 85.327 | 0.920 |
| Row ID | Condition | DL classifier | CV (%) | Mean of accuracies (%) | ASI |
| 26 | (*) clean | VGG19 | 3.980 | 91.129 | 0.916 |
| 27 | 5%_noise | VGG19_5P_SP0.1GA0.1 | 4.970 | 90.519 | 0.896 |
| 28 | 10%_noise | VGG19_10P_SP0.1GA0.1 | 5.867 | 89.892 | 0.877 |
| 29 | 15%_noise | VGG19_15P_SP0.1GA0.1 | 3.042 | 90.988 | 0.935 |
| 30 | 5%_noise | VGG19_5P_GA0.1SP0.1 | 4.981 | 90.208 | 0.895 |
| 31 | 10%_noise | VGG19_10P_GA0.1SP0.1 | 4.922 | 90.306 | 0.897 |

| Row ID | Condition | DL classifier | CV (%) | Mean of accuracies (%) | ASI |
|---|---|---|---|---|---|
| 32 | 15%_noise | VGG19_15P_GA0.1SP0.1 | 5.327 | 89.662 | 0.888 |
| 33 | 5%_noise | VGG19_5P_SP0.1RR30 | 3.165 | 90.862 | 0.933 |
| 34 | 10%_noise | VGG19_10P_SP0.1RR30 | 3.089 | 90.991 | 0.934 |
| 35 | 15%_noise | VGG19_15P_SP0.1RR30 | 2.718 | 92.009 | 0.943 |
| 36 | 5%_noise | VGG19_5P_SP0.1RL30 | 2.761 | 91.215 | 0.941 |
| 37 | 10%_noise | VGG19_10P_SP0.1RL30 | 3.198 | 91.527 | 0.932 |
| 38 | 15%_noise | VGG19_15P_SP0.1RL30 | 1.945 | 91.564 | 0.958 |
| 39 | 5%_noise | VGG19_5P_SP0.1GA0 | 5.600 | 89.933 | 0.883 |
| 40 | 10%_noise | VGG19_10P_SP0.1GA0 | 2.841 | 91.334 | 0.940 |
| 41 | 15%_noise | VGG19_15P_SP0.1GA0 | 2.940 | 91.108 | 0.937 |
| 42 | 5%_noise | VGG19_5P_GA0.1SP0 | 4.194 | 90.958 | 0.912 |
| 43 | 10%_noise | VGG19_10P_GA0.1SP0 | 4.422 | 90.361 | 0.907 |
| 44 | 15%_noise | VGG19_15P_GA0.1SP0 | 5.389 | 89.940 | 0.887 |
| 45 | 5%_noise | VGG19_5P_RL30SP0 | 2.511 | 92.262 | 0.947 |
| 46 | 10%_noise | VGG19_10P_RL30SP0 | 2.720 | 91.599 | 0.942 |
| 47 | 15%_noise | VGG19_15P_RL30SP0 | 2.243 | 92.017 | 0.952 |
| 48 | 5%_noise | VGG19_5P_RR30SP0 | 2.824 | 91.706 | 0.940 |
| 49 | 10%_noise | VGG19_10P_RR30SP0 | 2.454 | 92.125 | 0.948 |
| 50 | 15%_noise | VGG19_15P_RR30SP0 | 2.523 | 91.951 | 0.947 |
| Row ID | Condition | DL classifier | CV (%) | Mean of accuracies (%) | ASI |
| 51 | (*) clean | ResNet | 2.556 | 88.558 | 0.944 |
| 52 | 5%_noise | ResNet50_5P_SP0.1GA0.1 | 1.951 | 88.688 | 0.957 |
| 53 | 10%_noise | ResNet50_10P_SP0.1GA0.1 | 2.262 | 88.620 | 0.950 |
| 54 | 15%_noise | ResNet50_15P_SP0.1GA0.1 | 3.181 | 87.617 | 0.930 |
| 55 | 5%_noise | ResNet50_5P_GA0.1SP0.1 | 2.333 | 88.355 | 0.949 |
| 56 | 10%_noise | ResNet50_10P_GA0.1SP0.1 | 2.412 | 88.535 | 0.947 |
| 57 | 15%_noise | ResNet50_15P_GA0.1SP0.1 | 2.011 | 88.779 | 0.956 |
| 58 | 5%_noise | ResNet50_5P_SP0.1RR30 | 0.800 | 89.702 | 0.982 |
| 59 | 10%_noise | ResNet50_10P_SP0.1RR30 | 0.677 | 89.709 | 0.985 |
| 60 | 15%_noise | ResNet50_15P_SP0.1RR30 | 0.842 | 89.512 | 0.981 |
| 61 | 5%_noise | ResNet50_5P_SP0.1RL30 | 1.026 | 89.289 | 0.977 |
| 62 | 10%_noise | ResNet50_10P_SP0.1RL30 | 0.958 | 89.759 | 0.979 |
| 63 | 15%_noise | ResNet50_15P_SP0.1RL30 | 0.592 | 89.647 | 0.987 |
| 64 | 5%_noise | ResNet50_5P_SP0.1GA0 | 2.452 | 88.475 | 0.946 |
| 65 | 10%_noise | ResNet50_10P_SP0.1GA0 | 1.936 | 88.831 | 0.957 |
| 66 | 15%_noise | ResNet50_15P_SP0.1GA0 | 2.209 | 88.410 | 0.951 |
| 67 | 5%_noise | ResNet50_5P_GA0.1SP0.1 | 2.515 | 88.515 | 0.945 |
| 68 | 10%_noise | ResNet50_10P_GA0SP0.1 | 2.184 | 88.625 | 0.952 |
| 69 | 15%_noise | ResNet50_15P_GA0SP0.1 | 2.133 | 88.528 | 0.953 |
| 70 | 5%_noise | ResNet50_5P_RR30SP0 | 0.824 | 89.838 | 0.982 |
| 71 | 10%_noise | ResNet50_10P_RR30SP0 | 0.612 | 89.769 | 0.986 |
| 72 | 15%_noise | ResNet50_15P_RR30SP0 | 0.554 | 89.715 | 0.988 |
| 73 | 5%_noise | ResNet50_5P_RL30SP0 | 0.757 | 89.590 | 0.983 |
| 74 | 10%_noise | ResNet50_10P_RL30SP0 | 0.730 | 89.995 | 0.984 |
| 75 | 15%_noise | ResNet50_15P_RL30SP0 | 0.732 | 89.576 | 0.984 |